\documentclass[journal]{IEEEtran}
\usepackage{amsmath,amsfonts}
\usepackage{algorithmic}
\usepackage{algorithm}
\usepackage{array}
\usepackage[caption=false,font=normalsize,labelfont=sf,textfont=sf]{subfig}
\usepackage{textcomp}
\usepackage{stfloats}
\usepackage{url}
\usepackage{verbatim}
\usepackage{graphicx}
\usepackage{cite}
\usepackage{amssymb}
\usepackage{diagbox}
\usepackage{multirow}
\usepackage[table]{xcolor}
\usepackage{hhline}
\usepackage{threeparttable}
\usepackage{bbding}
\usepackage{makecell}
\usepackage{stackengine}

\begin{document}

\title{Dynamic Relevance Learning for Few-Shot Object Detection}

\author{Weijie Liu, Chong Wang*,~\IEEEmembership{Member,~IEEE}, Haohe Li, Shenghao Yu, Jiangbo Qian, Jun Wang and Jiafei Wu
\thanks{Manuscript received ***. This work was supported by the Zhejiang Provincial Natural Science Foundation of China (No. LY20F030005), Scientific Innovation 2030 Major Project for New Generation of AI, Ministry of Science and Technology of the People's Republic of China (No. 2020AAA0107300) and National Natural Science Foundation of China (No. 61603202). (Corresponding author: Chong Wang.)}
\thanks{W. Liu, C. Wang, H. Li and S. Yu  are with the Faculty of Electrical Engineering and Computer Science, Ningbo University, China (e-mail: liuweijie19980216@gmail.com; wangchong@nbu.edu.cn; lihaohe1023@163.com; ysh\_nbu@163.com).}
\thanks{J. Wu is with SenseTime Research, China (e-mail: wujiafei@sensetime.com).}}

\markboth{ }%
{Shell \MakeLowercase{\textit{W. Liu et al.}}: Dynamic Relevance Learning for Few-Shot Object Detection}

\maketitle

\begin{abstract}
Expensive bounding-box annotations have limited the development of object detection task. Thus, it is necessary to focus on more challenging task of few-shot object detection. It requires the detector to recognize objects of novel classes with only a few training samples. Nowadays, many existing popular methods adopting training way similar to meta-learning have achieved promising performance, such as Meta R-CNN series. However, support data is only used as the class attention to guide the detecting of query images each time. Their relevance to each other remains unexploited. Moreover, a lot of recent works treat the support data and query images as independent branch without considering the relationship between them. To address this issue, we propose a dynamic relevance learning model, which utilizes the relationship between all support images and Region of Interest (RoI) on the query images to construct a dynamic graph convolutional network (GCN). By adjusting the prediction distribution of the base detector using the output of this GCN, the proposed model serves as a hard auxiliary classification task, which guides the detector to improve the class representation implicitly. Comprehensive experiments have been conducted on Pascal VOC and MS-COCO dataset. The proposed model achieves the best overall performance, which shows its effectiveness of learning more generalized features. Our code is available at \url{https://github.com/liuweijie19980216/DRL-for-FSOD}.
\end{abstract}


\section{Introduction}
\IEEEPARstart{W}{ith} the rapid development of deep learning, computers have surpassed human beings in an increasing number of aspects, such as image classification \cite{ref1, ref2, ref3}. However, as far as current technologies are concerned, a large amount of labeled data must be provided so that machines can learn to adapt new tasks. In contrast, human can recognize a certain animal just by looking at a picture once. In order to further narrow the gap between the computer and human, how to make model learn a new task through only limited labeled samples has gradually become a hot research topic.

To address such challenging problem, meta-learning inspired models are widely used in few-shot learning with promising performance. Meta-learning could help model learn additional prior knowledge, to improve the generalization ability of the model. In this way, when the model encounters a new task, it can quickly learn to handle it. In other words, it makes the model learn to learn. Normally, meta-learning is based on four kinds of techniques, namely data augmentation \cite{ref4, ref5, ref6}, external memory \cite{ref7, ref8}, parameter optimization \cite{ref9, ref10, ref11} and metric learning \cite{ref12, ref13, ref14, ref15, ref16}.

\begin{figure}[!t]
\centering
\includegraphics[width=3.5in]{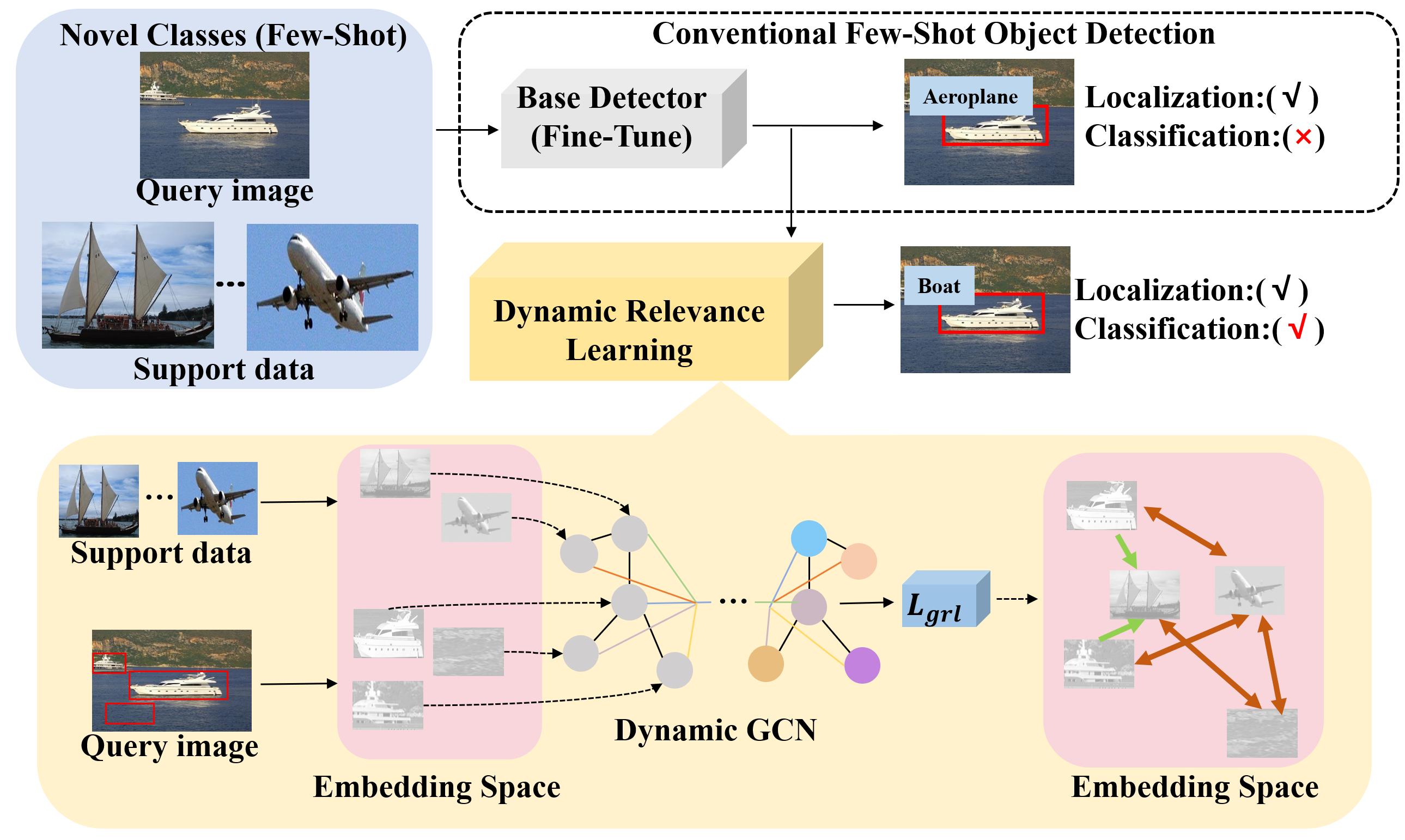}
\caption{Illustration of dynamic relevance learning (DRL). During the conventional fine-tune stage marked in the dashed block, the proposed DRL is added for boosting feature discriminability. The output of a dynamic GCN is used to form a new DRL loss, which penalize the close pairs of RoI and support features if they are not in the same category. }
\label{fig_1}
\end{figure}

At present, the few-shot learning has been widely used in computer vision, such as image and video classification \cite{ref17, ref18, ref19}, generation \cite{ref20}, translation \cite{ref21}, object detection \cite{ref22, ref23, ref24, ref25, ref26, ref27, ref28, ref29, ref30, ref31, ref32, ref33, ref34, ref35} and so on. Few-shot object detection (FSOD) requires the model to detect objects of novel classes only through a few support images. The existing methods \cite{ref22, ref23, ref24} adopt a parallel structure like meta-learning based models, which extracts features from support images and query images at the same time. The predictions of classification and regression are completed by using the features of the query images (or Region of Interest, RoI) combined with support features. Therefore, their relevance to each other becomes particularly important. However, some works only consider to increase the inter-class diversity in support data \cite{ref22}, or learn the metric to predict similarity between RoI features \cite{ref25}. The relationship between support and query features remains unexploited. 
  
In the conventional methods, the base model was first trained with sufficient samples of base classes. Then, such base model was fine-tuned with much less samples of novel classes, as shown in the dashed box in Fig. 1. Due to the limited number of samples, it is difficult to let the model learn the generalized features, which leads to erroneous classification. To better transform the knowledge of base model to novel classes, we exploit the relationship between support and RoI features in this paper to enhance the feature learning process. In other words, the dependency established between support data and RoI will make the model more discriminative. 

Meanwhile, deep metric learning aims to get more generalized features according to the relations of samples. Inspired by that, we propose a dynamic relevance learning (DRL) model by exploiting the similarities between features using a dynamic GCN. As illustrated in Fig. 1, the proposed model forms a new challenging classification task to train a better feature extractor. It is based on the basic assumption that similar features shall belong to the same class and share similar class probabilities. Otherwise, the performance of the feature extractor needs to be improved. Unlike other previous methods to use the feature similarity directly as a loss, the proposed DRL utilizes it implicitly to blend the class probabilities for a second classification. The complicate links between support and query features introduced by GCN can help the learning of feature representation, while DRL serves as an inspector to make sure the aforementioned assumption holds. The dynamic GCN designed in this work can not only update the class probabilities, but also capture the error of the initial probability vectors. By calculating the classification loss of the output nodes of the GCN, the features of the same category of support images and RoI will become more similar, and the features of different categories will be farther apart. 

The main contributions of this work are two-fold. 1) The proposed DRL opens up a new way to implicitly exploit the relevance between the support and query features. Unlike some previous works that manipulate the features directly, DRL utilizes feature correlation to construct a dynamic GCN and then forms a hard auxiliary classification task for better feature learning. 2) The input and output of the dynamic GCN is specifically designed to bridge the support and query set, under a new prior that similar features shall belong to the same class and share similar class probabilities. The labels and category probabilities, not the features, of the support images and RoIs in the query image are defined as anchor nodes and drift nodes, respectively. The output of GCN, i.e., enhanced category probabilities, is then applied to calculate the classification loss to guide the training of feature extractor.

\section{Related Works}
The dynamic relevance learning (DRL) proposed in this paper aims to improve the performance of few-shot object detection (FSOD), which is a subproblem in few-shot learning. The idea of deep metric learning is applied to constrain the distances between visual features, with the help of a dynamic GCN bridging the support and query images.

\begin{figure*}[!t]
\centering
\includegraphics[width=6.5in]{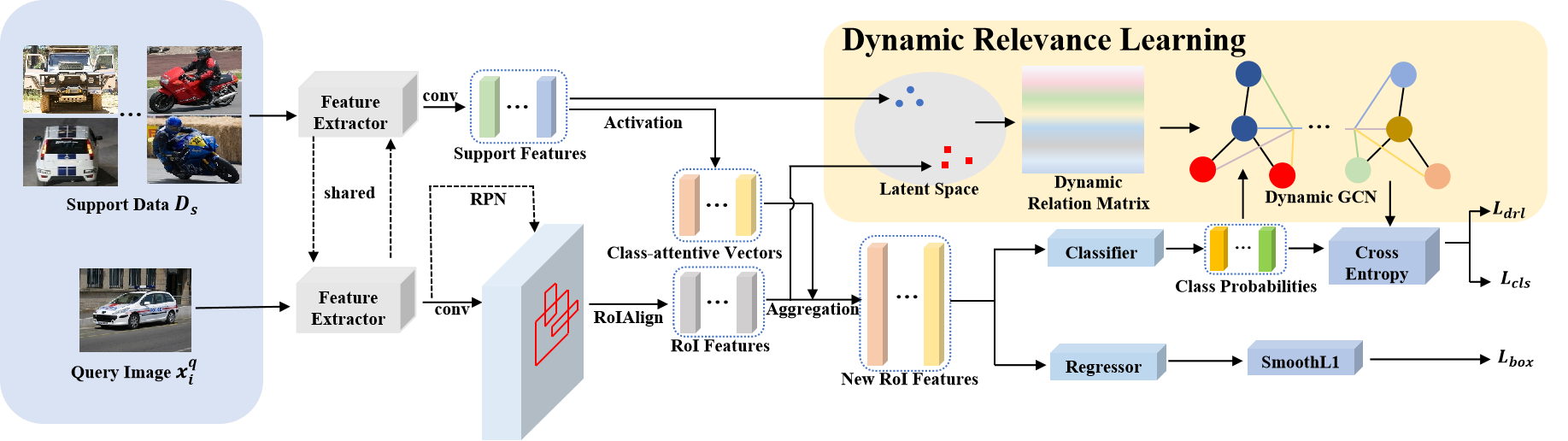}
\caption{Overview of the proposed model. It consists of a few-shot object detector and the dynamic relevance learning module.}
\label{fig_2}
\end{figure*}

\subsection{Few-Shot Learning}
Since deep learning based models require a large number of training samples, the models may fail to find the optimal parameters if the training data is insufficient. Most of few-shot learning algorithms introduce a suitable prior knowledge using meta-learning to narrow the space for model parameter adjustment with few samples \cite{ref36}. 

The most direct way to address the issue of insufficient training samples is to generate new ones, \cite{ref4, ref5, ref6} use data augmentation or Generative Adversarial Networks (GAN) to generate more samples. In order to better transfer knowledge to new tasks, \cite{ref7, ref8} use additional memory to save the previous information. For the images of novel classes, similar information will be searched in the memory to help identify them. Meanwhile, some methods based on parameter adjustment \cite{ref9, ref10, ref11} try to make the model learn to get an appropriate initialization parameter for specific tasks. Metric learning based methods \cite{ref12, ref13, ref14, ref15, ref16} calculate the similarity between query and support images of each category to get the predicted category probability. Sharing a similar motivation of this work, both TRPN \cite{ref37} and  BatchFormer \cite{ref38} explore sample relationships by Transformer Encoder and GCN, to help model learn novel knowledge with a few samples. However, the way of learning feature similarities in the proposed DRL are different, which focus on implicitly guiding the learning process by adjusting the category probability instead of manipulating visual features directly.

\subsection{Few-Shot Object Detection}
Object detection is a challenging problem in computer vision. At present, the deep learning models\cite{ref39} have gradually replaced the traditional machine learning methods, and become the mainstream algorithms in the field of object detection. Most of the algorithms can be divided into two categories, namely one-stage models \cite{ref40, ref41, ref42} and two-stage models \cite{ref43, ref44, ref45, ref46}, by whether there is a separated process of region proposals.

In recent years, how to use a small number of samples for object detection has become an active topic. RepMet \cite{ref27} is the first framework for FSOD, which simply replaces the classification head of the traditional two-stage detection algorithm with the classification method of Matching Network \cite{ref14}. LSTD \cite{ref28} proposes a framework transferring knowledge that the model has learned to the novel classes. MetaYOLO \cite{ref26} uses the information of support images to get the class-attentive vectors, which guides the query images to complete the detection task. MetaDet \cite{ref29} decomposes the parameters of the detector into two parts, namely category-agnostic and category-specific. Meta R-CNN \cite{ref22} combines class-attentive vectors with RoI feature instead of the whole image feature to make the object recognition more accurately.

TFA \cite{ref30} proposes a simple method based on a fine-tune process, which achieves promising results in the case of few samples. Then, FSCE \cite{ref31} introduces the Contrastive Learning into TFA framework, by adding a specific contrastive head responsible for enhancing the feature discrimination of novel classes. Meanwhile, Retentive R-CNN \cite{ref32} designs the trainable sub-networks for identifying novel classes in RPN and detector respectively, which can improve the performance of novel classes and maintain the performance of base classes as much as possible. And SRR-FSD \cite{ref33} utilizes sematic information to make detection results more stable. Recently, N-PME \cite{ref34} proposes a new strategy to mine pseudo novel instances to finetune the detector. The detector used in this paper is also based on the FSOD framework like Meta R-CNN \cite{ref22}, but a dynamic relevance learning model is proposed to establish a stronger link between support and query (RoI) features.

\subsection{Deep Metric Learning}
Many deep metric learning (DML) methods have been widely used in the field of FSOD, such as \cite{ref27}. DML aims to utilize the neural networks to obtain features with strong discrimination through a simple metric function constraint. In the same embedding space, the distance between samples of the same categories becomes smaller, while the distance between samples of different categories is enlarged. Contrastive loss \cite{ref47} introduces metric learning into deep neural network for the first time. To be specific, samples are paired in pairs to reduce the distance between samples within the same category and maintain a certain distance for different categories. Triplet loss {ref48} further considers the relationship between inter-class pairs as well as intra-class pairs. Meanwhile, many latest works focus on improving the sampling strategy \cite{ref49, ref50} for DML. Especially, the proxy-based methods \cite{ref51, ref52, ref53} have attracted extensive attention due to its capability to alleviate the difficulty of sampling. 

Recently, group loss \cite{ref53} has been proposed to take some samples of mini-batch as anchors and constrain their distances with other samples. The class probabilities are iteratively updated in an empirical way to exploit the similarities between features. However, such method may introduce undesirable bias, if the initial probability distribution is inaccurate which is often true unfortunately. The error will be accumulated after several iteration and lead to worse performance. To address this issue, we propose a more sophisticated model using a dynamic GCN to process the probability distribution from different categories.

\subsection{Graph Convolutional Network}
The graph convolutional neural network (GCN) plays a key role in the proposed DRL model. Traditional CNN is good at processing Euclidean spatial data, but a lot of data is in the form of graph data in real life. Migrating CNN to analysis and process graph data, GCN was first proposed in \cite{ref54} to address the semi-supervised classification problem. Subsequently, many improvements of GCN have been proposed \cite{ref55, ref56}. 

Recently, there is a new trend that utilizing GCN to help models in recognition tasks. To obtain better representation of video content, a Dual-Pooling GNN is proposed in \cite{ref17}. In Zero-shot learning, GCN is used in \cite{ref57} with semantic embeddings and categorical relationships of knowledge graph to predict the classifiers. To improve the categorical relationships, a Dense Graph Propagation (DGP) module is proposed in\cite{ref58}. Furthermore, a novel ML-ZSL approach \cite{ref59} is proposed to explore the associated structured knowledge. In \cite{ref60}, GCN is introduced to form the attribute propagation network (APNet). In zero-shot detection, GCN has also been used in Semantics-Preserving Graph Propagation model (SPGP) \cite{ref61}. It can effectively use the semantic embeddings and structural knowledge given in the previous category diagram to enhance the generalization ability of the learned projection function. 

Most of the aforementioned works generate a static graph for GCN in advance based the relationships between categories. However, due to the changing of data in some cases, different graph structures may need to be established during the training. In such case, dynamic GCN \cite{ref62} is required. Unfortunately, very few related researches have been reported. In this paper, a new type of dynamic GCN is proposed for deep metric learning.

\section{Methodology}\
In this section, the setup for few-shot object detection is first introduced, followed by the general description of the proposed framework. Then, the dynamic relevance learning (DRL) algorithm is discussed in details. The training strategy for the proposed DRL is presented at the end.

\subsection{Problem Setup}
In the task of Few-Shot Object Detection (FSOD), the dataset is usually divided into two parts: base classes with $ N_b $ samples and novel classes with $ N_n $ samples, where $ N_b \gg N_n $. The corresponding data of these two parts are denoted as $ D_b $ and $ D_n $. The goal of FSOD is to train a detector $ h(;\theta) $ on $ D_b $ and quickly adapt it to $ D_n $, where $ \theta $ is the learnable parameters. In this work, a episodic-training strategy is used in FSOD, which has been widely used for few-shot learning \cite{ref22, ref23, ref24}. As shown in Fig. 2, each mini-batch consists of a support set $ D_s = \{(x_{c, k}^s, y_{c, k}^s)\}_{c=1; k=1}^{N; K} $ and a query set  $ D_q = \{(x_{i}^q , y_{i}^q)\}_{i=1}^{n_q} $. There are $ n_q $ query images in $ D_q $, $ N $ categories and $ K $ images in $ D_s $ respectively, which is called $ N $-way $ K $-shot. In this work, full-way $ K $-shot is used, i.e., each support set contains images of all categories in the dataset. Since the data in the support set is labeled with $ y_{c, k}^s $, the model could learn to adaptively adjust the prediction results of the query set in an episodic-training manner. 

\subsection{Overview}
As shown in Fig. 2, our framework is mainly divided into two parts, the detector based on meta-learner $ h(; D_s; \theta) $ and the Dynamic Relevance Learning responsible for associated learning between support features and query features. The detector utilizes convolutional neural networks (CNNs) as the feature extractor to simultaneously generate the feature maps from the support data $ D_s $ and query $ i $-th image $ x_i^q$ . Then, the Region Proposal Network (RPN) and RoIAlign are used to obtain the RoI features $ \{r_{i,j}\}_{j=1}^{n_{roi}} $ from the query feature maps. The average value of $ K $ sample features in support data $ D_s $ corresponding to class $ c $ is used as the class representation. Then the class-attentive vector $ a_c $ is obtained as, 

\begin{figure*}[!t]
\centering
\includegraphics[width=6.5in]{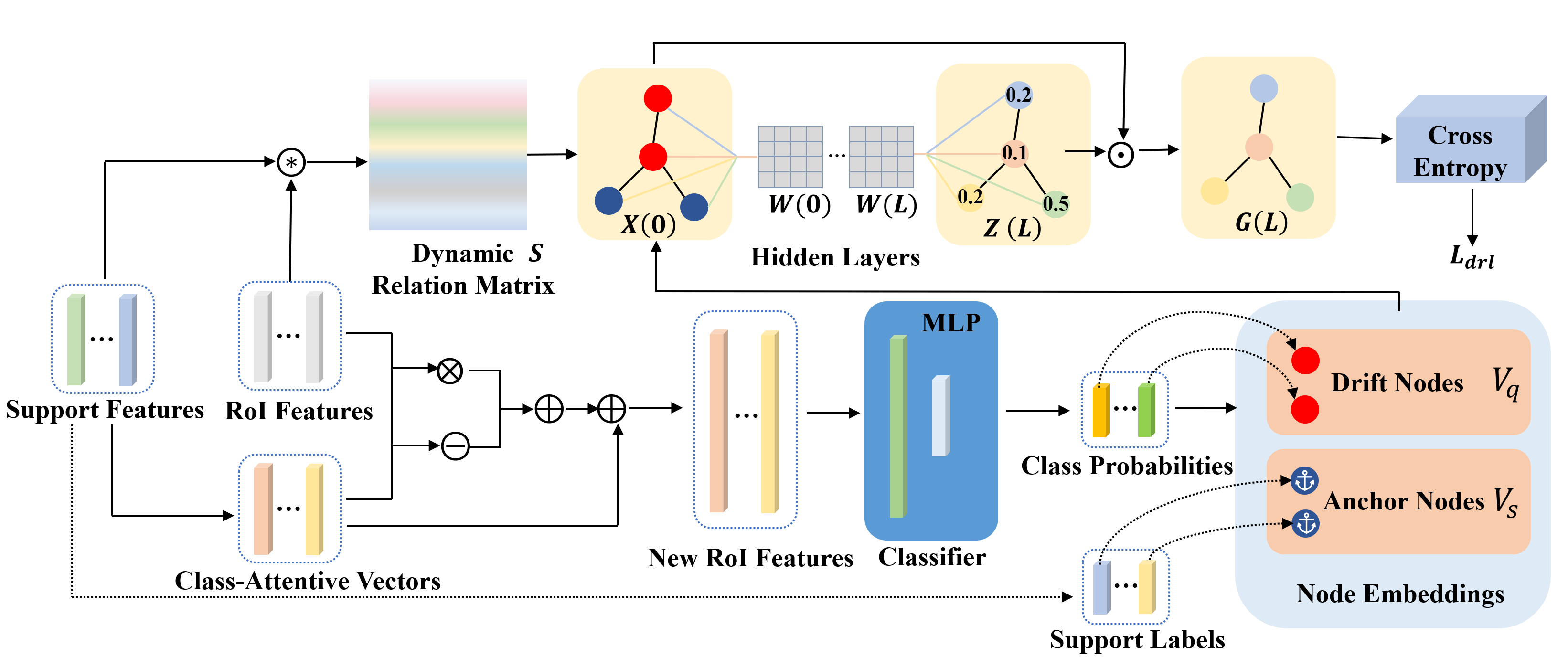}
\caption{Dynamic relevance learning. $ \otimes $ and $ \ominus $ represent channel-wise multiplication and subtraction, respectively. $ \oplus $ denotes concatenation operation, $ \circledast $ calculates the Pearson’s correlation coefficient, and $ \odot $ is node-wise multiplication. }
\label{fig_3}
\end{figure*}

\begin{align}
\label{eq1,2}
v_{c, k} = F^{exr} (x_{c, k}^s), \\
a_c = \delta (\frac{1}{K}\sum_{k=1}^K v_{c, k}),
\end{align}
where $F^{exr}(*) $ denotes the feature extractor (i.e. CNNs), $ v_{c,k} $ is the support features of the $ k $-th image in $ c $-th category and $ \delta(*) $ presents the sigmoid activation function.

The class-attentive vector $ a_c $ and the RoI feature $ r_{i,j} $  are then aggregated and fed into the prediction head. The aggregation operation $ \mathcal{A} $ proposed in [24] is adopted in our model to obtain the new RoI feature $ \hat{r}_{i, j} $ ,

\begin{equation}
\begin{split}
\label{eq3}
\hat{r}_{i, j} &= \mathcal{A}(a_c, r_{i,j})\\ 
&= [r_{i,j} \otimes a_c; r_{i,j} \ominus a_c; r_{i,j}],
\end{split}
\end{equation}
where $ \otimes $, $ \ominus $ and $ [ ; ] $ represent channel-wise multiplication, subtraction and concatenate operation, respectively. Then the classification probability distribution $ \{p_{i,j}\}_{j=1}^{n_{roi}} $ is calculated by the classification head through the new RoI features $ \{\hat{r}_{i,j}\}_{j=1}^{n_{roi}} $. 

\begin{equation}
\label{eq4}
p_{i,j} = softmax(F^{cls}(\hat{r}_{i, j})), p_{i,j} \in \mathbb{R}^C, 
\end{equation}
where $ F^{cls}(*) $ represents the classification head function which is a Multi-Layer Perception (MLP) in this work. At the end, the classification loss $ L_{cls} $ and regression loss $ L_{reg} $ can be obtained through the $ \{p_{i,j}\}_{j=1}^{n_{roi}} $, $ \{\hat{r}_{i,j}\}_{j=1}^{n_{roi}} $ and the labels.
 
Although the meta-learning inspired structure can alleviate the data imbalance problem between the $ C_b $ and $ C_n $, the detection performance on $ D_n $ is still unsatisfactory. The main reason is that the limited support samples from $ D_n $ lead to unstable class representation. Moreover, such unstable representation may introduce confused information in the RoI learning stage, since $ v_{c,k} $ is aggregated into the RoI feature $ \hat{r}_{i,j} $. 

To address this issue, we propose a Dynamic Relevance Learning paradigm. The key concept is using the dependency (between support and RoI features) modeled by a dynamic GCN to improve the class representation implicitly. Specifically, the support features and RoI features are mapped to a latent space, while their similarity matrix is adopted as the adjacency matrix of dynamic GCN. The previous class probabilities $ \{p_{i,j}\}_{j=1}^{n_{roi}} $ and the support data labels $ \{y_{c,k}^s\}_{c=1; k=1}^{C; K} $ are utilized as the node embeddings in the graph convolution. This dynamic GCN is further constrained by the classification loss $ L_{drl} $ of the node embeddings during the training stage. The details of this dynamic relevance learning will be discussed in the next section.

\subsection{Dynamic Relevance Learning}
When transferring the base model trained on $ D_b $ with sufficient samples to $ D_n $ with only limited samples, it often suffers a great performance degradation. This is obviously due to the lack of samples that makes the model struggling to learn the desirable feature representation. In most frameworks of meta-learning, the support data $ D_s $ provide general class-attentive vectors, which are equivalent to the category templates for RoI learning. To further explore the relation between the support and RoI features, a dynamic relevance learning paradigm is proposed to build proper connections between them. Two types of dynamic GCN are designed to adjust the prediction classification probability distribution, which guides the feature representation learning implicitly. Each node embedding in GCN is updated according to the nodes connected with it. Thus, it is naturally suitable to model the relevance between categories. 

Given a graph $ G = \{V,E,A\} $ for constructing GCN, $ V $ and $ E $ are the sets of nodes and edges respectively and $ A $ is the adjacency matrix indicating the relationship between each node. In our task, the nodes $ V = \{V_s, V_q\} $ represent either the support images $ (V_s) $ or RoIs $ (V_q) $ from the query image $ x_i^q $. As shown in Fig. 3, the similarity matrix  $ S $ of support features  $ \{v_{c,k}\}_{c=1; k=1}^{C; K} $ and RoI features $ \{r_{i,j}\}_{j=1}^{n_{roi}} $  is adopted as the adjacency matrix of graph $ G $, which is calculated using Pearson’s correlation coefficient,

\begin{equation}
\label{eq5}
s_{m,n} = \frac{Cov(f_m, f_n)}{\sqrt{Var(f_m, f_n)}},
\end{equation}
where $ s_{m,n} $ is the element of matrix $ S \in \mathbb{R}^{M \times M} $, $ M = CK + n_{roi} $ and $ f_m $, $ f_n $ are features sampled from the union of $ \{v_{c,k}\}_{c=1; k=1}^{C; K} $ and $ \{r_{i,j}\}_{j=1}^{n_{roi}} $. It is worth noting that, in conventional GCN, the graph $ G $ is determined before the training stage and such graph structure will not be altered throughout the whole training process. However, the support data $ D_s $ and the query image $ x_i^q $ changes at each training iteration. In other words, the graph $ G $ keeps changing during the training process. Thus, instead of building a static graph in advance, the dynamic GCN is employed to learning the dynamic relevance between the altering nodes, while matrix $ S $ is now viewed as a dynamic relation matrix. 

Inspired by group loss \cite{ref53}, a small portion of samples are selected as anchor nodes in each training mini-batch, while the rest are denoted as drift nodes. The output of these anchor nodes is identical to their input node embeddings, i.e., the labels, which provide a strong guidance for the drift nodes. Specifically, as shown in Fig. 3, the support images $ (V_s) $ are chosen as anchor nodes, while the one-hot coded support labels $ y_{c,k}^s $ and the RoI classification probabilities $ p_{i,j} $  are considered as the node embeddings for $ V_s $ and $ V_q $, respectively. Since the support data are well-labeled, they can produce a reliable guiding effect for RoI feature learning. To be noted, the RoI features $ (V_q) $ generated by RPN are not suitable to be the anchor nodes, because either their position or category classification are inaccurate.

\begin{figure}[!t]
\centering
\subfloat[]{\includegraphics[width=1in]{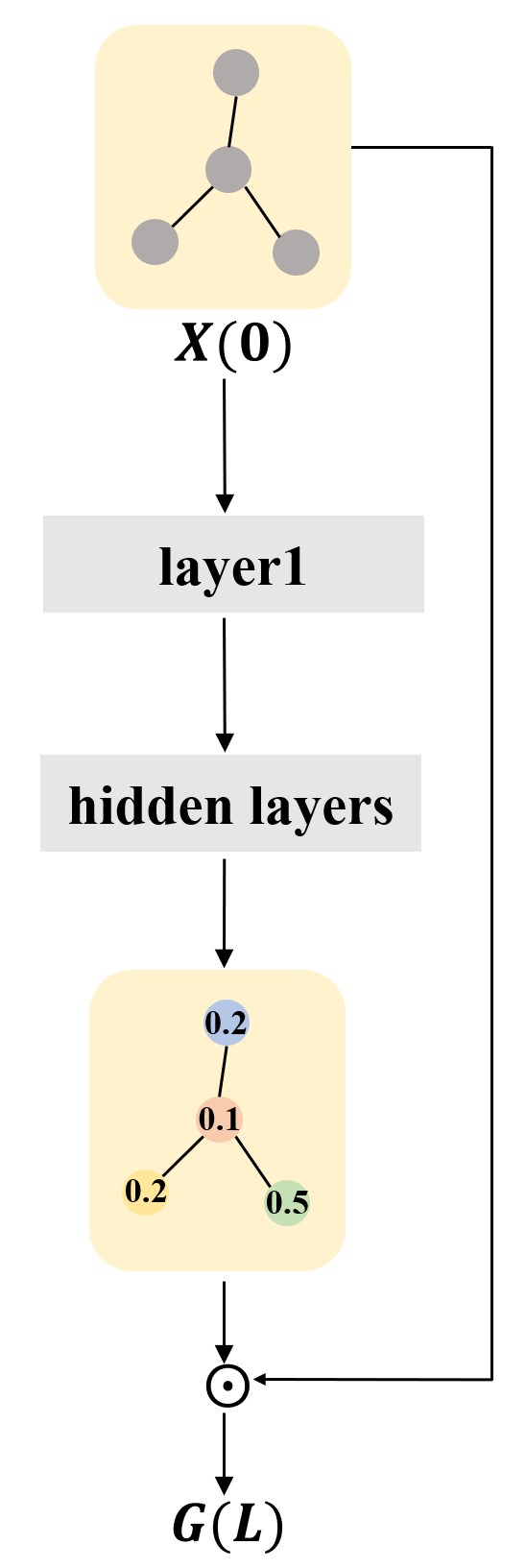}%
\label{fig_normal}}
\hfil
\subfloat[]{\includegraphics[width=1in]{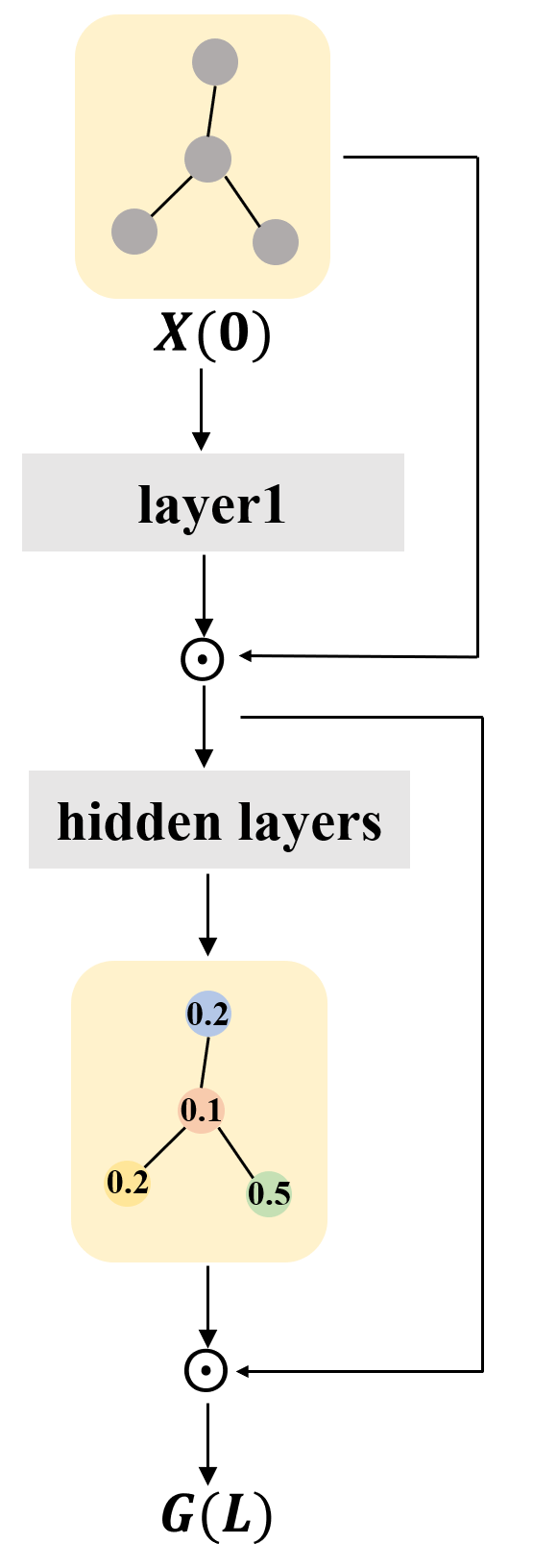}%
\label{fig_residual}}
\caption{Two structures of dynamic GCN. (a) Normal structure. (b) Residual structure. $ \odot $ Represents the multiplication of the corresponding nodes.}
\label{fig_4}
\end{figure}

According to the dynamic relation matrix $ S $, the anchor and drift nodes and their corresponding node embeddings, the input $ X(0) $ of the proposed dynamic GCN is, 

\begin{equation}
\label{eq6}
X(0) = \{p_{i,j}\}_{j=1}^{n_{roi}} \cup \{y_{c,k}^s\}_{c=1; k=1}^{C; K}.
\end{equation}

Since the relation matrix $ S $ defines a complete graph, its diagonal degree matrix is actually an identity matrix multiplied by the number of nodes. Thus, the graph convolution in this work is simplified as, 

\begin{equation}
\label{eq7}
Z(l) = \sigma(SX(l)W(l)),
\end{equation}
where $ l = 1,2,3,...,L $ is the layer index of GCN, $ X(l) $ is input of layer $ l $, $ W(l) \in \mathbb{R}^{C \times C} $ is the learnable weight matrix of layer $ l $, $ \sigma $ is a non-linear activation function, and $ Z(l) $ is the output of layer $ l $. 

The output $ Z(L) $ of the $ L $-th layer is considered as the relevance attention to indicate the confidence on each category according to the input node embedding and connected nodes. As shown in Fig. 4(a), these weights are multiplied back to the initial input $ X(0) $ to get the updated node embeddings $ G(L) $,

\begin{equation}
\label{eq8}
G(l) = Z(l) \odot X(0).
\end{equation}

\begin{table*}[!t]
\renewcommand\arraystretch{1.8}
\caption{Evaluation Results on PASCAL VOC 2007 Test Set ($ AP_{50} $, \%)\label{tab:table1}}
\centering
\setlength\doublerulesep{0.5pt}
\begin{threeparttable}
\begin{tabular}{c|ccccc|ccccc|ccccc}
\hhline{================}
\multirow{2}{*}{\diagbox{Methods}{Shots}}&\multicolumn{5}{c|}{Novel Set 1}&\multicolumn{5}{c|}{Novel Set 2} &\multicolumn{5}{c}{Novel Set 3}\cr & 1 & 2 & 3 & 5 & 10 & 1 & 2 & 3 & 5 & 10 & 1 & 2 & 3 & 5 & 10 \cr\hline
LSTD \cite{ref28}  & 8.2 & 1.0 & 12.4 & 29.1 & 38.5 & 11.4 & 3.8 & 5.0 & 15.7 & 31.0 & 12.6 & 8.5 & 15.0 & 27.3 & 36.3 \cr

MetaYOLO \cite{ref26} & 14.8	& 15.5 &26.7 & 33.9 & 47.2	& 15.7 & 15.2 & 22.7	& 30.1 & 40.5 & 21.3 & 25.6 & 28.4 & 42.8	& 45.9\cr
MetaDet \cite{ref29}	& 18.9	& 20.6	& 30.2	& 36.8	& 49.6	& 21.8	& 23.1	& 27.8	& 31.7	& 43.0	& 20.6	& 23.9	& 29.4	& 43.9	& 44.1\cr
Meta R-CNN \cite{ref22}	& 19.9	& 25.5	& 35.0	& 45.7	& 51.5	& 10.4	& 19.4	& 29.6	& 34.8	& 45.4	& 14.3	& 18.2	& 27.5	& 41.2	& 48.1\cr
TFA \cite{ref30} 	& 25.3	& 36.4	& 42.1	& 47.9	& 52.8	& 18.3	& 27.5	& 30.9	& 34.1	& 39.5	& 17.9	& 27.2	& 34.3	& 40.8	& 45.6 \cr
FSDVE \cite{ref24}	& 24.2	& 35.3	& 42.2	& 49.1	& 57.4	& 21.6	& 24.6	& 31.9	& \textbf{37.0}	& 45.7	& 21.2	& 30.0	 & 37.2  & 43.8 	& 49.6 \cr
\rowcolor{gray!30}
DRL (normal)$^{\ast}$       	& \textbf{30.3}	& \textbf{40.8}	& 49.1	& 48.0	& 58.6	& 22.4	& \textbf{36.1}	& \textbf{36.9}	& 35.4	& 51.8	& 24.8	& 29.3	& 37.9	& 43.6	& 50.4 \cr
\rowcolor{gray!30}
DRL (residual)$^{\ast}$	& 28.0	& 40.5	& \textbf{49.4}	& \textbf{49.9}	& \textbf{59.4}	& \textbf{22.9}	& 33.4	& 36.4	& 36.1	& \textbf{52.7}	& \textbf{28.0}	& \textbf{32.0}	& \textbf{40.4}	& \textbf{46.7}	& \textbf{53.5} \cr
\hhline{================}
\end{tabular}

\begin{tablenotes}
\footnotesize 
\item $ ^{\ast} $   The reported results are the average of ten random runs.
\end{tablenotes}
\end{threeparttable}

\end{table*}

\begin{table}[!t]
\renewcommand\arraystretch{1.2}
\caption{AP of Different Classes on VOC Novel Set 2\label{tab:table2}}
\centering
\setlength\doublerulesep{0.5pt}
\begin{threeparttable}
\begin{tabular}{c|c|ccccc}
\hhline{=======}
\multirow{2}{*}{Shots} &\multirow{2}{*}{Methods}  &\multicolumn{5}{c}{Classes} \cr 
& &aeroplane &bottle	&cow	&horse	&sofa \cr\hline

1 & FSDVE$ ^{\ast} $\cite{ref24} &35.8	&0.3	&\textbf{31.0}	&10.6	&\textbf{30.0} \cr &DRL (residual)	&\textbf{36.8}	&\textbf{8.5}	&25.3	&\textbf{15.1}	&28.8 \cr\hline

3 & FSDVE$ ^{\ast} $\cite{ref24} &\textbf{47.6}	&0.4	&48.9	&29.5	&44.1 \cr &DRL (residual)	&44.5	&\textbf{5.7}	&\textbf{49.3}	&\textbf{35.5}	&\textbf{46.7} \cr\hline

5 & FSDVE$ ^{\ast} $\cite{ref24} &\textbf{50.5}	&0.9	&\textbf{53.9}	&\textbf{41.0}	&41.5 \cr &DRL (residual) &43.9	&\textbf{2.9}	&50.2	&35.0	&\textbf{48.4} \cr\hline

10 & FSDVE$ ^{\ast} $\cite{ref24} &\textbf{55.8}	&15.1	&56.7	&66.1	&50.2 \cr &DRL (residual) &52.9	&\textbf{21.1}	&\textbf{63.9}	&\textbf{72.1}	&\textbf{53.6} \cr
\hhline{=======}
\end{tabular}

\begin{tablenotes}
\footnotesize 
\item $ ^{\ast} $   The reported results are reproduced using the official code. The mAPs are slightly higher than the ones reported in their paper \cite{ref24} shown in Table I.
\end{tablenotes}
\end{threeparttable}

\end{table}

In such manner, the original probability distribution $ \{p_{i,j}\}_{j=1}^{n_{roi}} $ will be modified based on the relevance attention $ Z(L)$ , which is determined by the relation between RoI and support features. This structure is slightly different from the traditional GCN, since we found it performs better to learn the attention scores instead of a modified probability distribution directly.

To consider the over-smoothing issue in GCN, we propose the other residual-like structure of GCN, as shown in Fig. 4(b). Instead of directly using  $ Z(l) $ (the output of $ l $-th layer) as the input of next layer, $ X(l+1) $ is obtain by multiplying $ Z(l) $ by the input of $ l $-th layer $ X(l) $ as,

\begin{equation}
\label{eq9}
X(l+1) = Z(l) \odot X(l).
\end{equation}

According to the experimental results, the performance of residual structure is generally better than the normal structure if there are adequate samples. More detailed discussion will be given in Section IV.

The output $ G(L) $ of GCN as the enhanced probability is then used to calculate the loss through Cross Entropy function. Formally, the loss $ L_{drl} $ is defined as,

\begin{equation}
\label{eq10}
L_{drl} = -\frac{1}{n_{roi}} \sum_{j=1}^{n_{roi}}y_j^q \log g_j(L),
\end{equation}
where $ g_j(L) $ is $ j $-th 	node embedding of $ G(L) $.

This loss adds implicit constraints between the RoI and support features. If a certain pair of RoI and support features shares a high similarity, strong link will be made between these two kinds of nodes (anchor and drift nodes). As a result, the drift node tends to give a high confidence at the same category the anchor node belongs to, which makes the predicted probability distribution close to the support label. If this prediction is correct, $ L_{drl} $ will be small. Otherwise, it gives penalty on such wrong relevance, which will encourage the model to increase the gap between them in the visual feature space.

\subsection{Training strategy}
 
The training process can be divided into two stages. The first stage uses a large amount of base class data to train the model, and the second one adopts a small amount of mixed data of novel and base classes. Due to the large difference in the data amount between these two stages, different loss functions are employed. Formally, the loss of base training $ L_{base} $ is defined as,

\begin{equation}
\label{eq11}
L_{base} = L_{rpn} + L_{cls} + L_{box} + L_{meta} + L_{drl},
\end{equation}
where $ L_{rpn} $ is proposed in Faster-RCNN for training RPN with better proposals. $ L_{cls} $ and $ L_{box} $ are proposed in Fast-RCNN to train the classification and regression head. $ L_{meta} $ is proposed in Meta R-CNN to get more stable support representation. $ L_{drl} $ is proposed in last section to establish proper relevance of support features and RoI features.

The meta loss $ L_{meta} $ has a relatively simple form designed to diversify the inferred object attentive vectors. It works well if there are sufficient training samples. However, it may cause the support features deviate from the true representation, if only few samples are available in the fine-tune phase. Therefore, the meta loss $ L_{meta} $ is not applied in the fine-tune phase. Formally, the loss of fine-tune phase $ L_{ft} $ is defined as,

\begin{equation}
\label{eq12}
L_{ft} = L_{rpn} + L_{cls} + L_{box} + L_{drl}.
\end{equation}

In addition, removing $ L_{drl} $ from (\ref{eq11}) in the base training stage will not affect the performance on  $ D_n $ too much. In the inference stage, we simply use the original probability distribution $ \{p_{i,j}\}_{j=1}^{n_{roi}} $ for classification. Even without the dynamic relevance learning part, the performance is still improved due to better feature representation learned in the training stage.

\section{Experiments}
In this section, the experimental results on Pascal VOC \cite{ref63} and MS-COCO \cite{ref64} dataset of the proposed dynamic relevance learning are presented, which are also compared with the state-of-the-art methods. Ablation studies and detailed analysis are also included.

\begin{table*}[!t]
\renewcommand\arraystretch{1.5}
\caption{Evaluation Results on MS-COCO 2014 Dataset (\%)\label{tab:table3}}
\centering
\setlength\doublerulesep{0.5pt}
\begin{threeparttable}
\begin{tabular}{c|c|cccccc|cccccc}
\hhline{==============}
Shots &Methods &$ AP $  &$ AP_{50} $  &$ AP_{75} $  &$ AP_S $ &$ AP_M $ &$ AP_L $ &$ AR_1 $ &$ AR_{10} $ &$ AR_{100} $ &$ AR_S $ &$ AR_M $ &$ AR_L $ \cr\hline

\multirow{10}{*}{10} &Meta R-CNN \cite{ref22} &8.7	&19.1	&6.6	&2.3	&7.7	&14.0	&12.6	&17.8	&17.9	&7.8	&15.6	&27.2 \cr
&Meta-RCNN \cite{ref23}	&9.4	&17.1	&9.4	&1.7	&11.2	&\textbf{18.1}	&--	&--	&--	&--	&--	&-- \cr
&TFA \cite{ref30} 	&9.1	&17.1	&8.8	&--	&--	&--	&--	&--	&--	&--	&--	&-- \cr
&FSDVE$ ^\ddag $ \cite{ref24}	&10.5	&\underline{25.5}	&5.7	&\underline{4.0}	&\underline{11.4}	&14.7	&18.6	&23.8	&23.9	&\underline{8.8}	&25.4	&32.2 \cr
&FSCE \cite{ref31}	&11.1	&--	&\underline{9.8}	&--	&--	&--	&--	&--	&--	&--	&--	&-- \cr
&Retentive R-CNN \cite{ref32} 	&10.5	&--	&--	&--	&--	&--	&--	&--	&--	&--	&--	&-- \cr
&SRR-FSD \cite{ref33} 	&\underline{11.3}	&23.0	&\textbf{9.8}	&--	&--	&--	&--	&--	&--	&--	&--	&-- \cr
&N-PME \cite{ref34} 	&10.6	&21.1	&9.4	&\textbf{4.6}	&9.4	&16.8	&16.4	&\textbf{27.6}	&\textbf{28.6}	&\textbf{13.3}	&\underline{26.2}	&\textbf{41.8} \cr
\rowcolor{gray!30}
&DRL (normal)$ ^\ast $       	&\textbf{11.9}	&\textbf{27.4}	&7.9	&3.8	&\textbf{12.6}	&\underline{17.7}	&\textbf{19.7}	&\underline{25.2}	&\underline{25.3}	&8.6	&\textbf{26.7}	&\underline{35.5} \cr
\rowcolor{gray!30}
&DRL (residual)$ ^\ast $	&10.9	&25.2	&7.0	&3.6	&11.2	&16.0	&\underline{19.0}	&24.7	&24.7	&8.1	&25.7	&34.4 \cr\hline

\multirow{10}{*}{30} &Meta R-CNN \cite{ref22} &12.4	&25.3	&10.8	&2.8	&11.6	&19.0	&15.0	&21.4	&21.7	&8.6	&20.0	&32.1 \cr
&Meta-RCNN \cite{ref23}	&12.8	&25.5	&12.2	&2.3	&12.3	&19.3	&--	&--	&--	&--	&--	&-- \cr
&TFA \cite{ref30} 	&12.1	&22.0	&12.0	&--	&--	&--	&--	&--	&--	&--	&--	&-- \cr
&FSDVE$ ^\ddag $ \cite{ref24}	&14.6 	&31.2	&11.4 	&\underline{5.3}	&15.3	&21.8 	&\underline{22.3} 	&29.3 	&29.5 	&\underline{11.4} 	&\textbf{31.4} 	&39.1 \cr
&FSCE \cite{ref31}	&\textbf{15.3}	&--	&\textbf{14.2}	&--	&--	&--	&--	&--	&--	&--	&--	&-- \cr
&Retentive R-CNN \cite{ref32} 	&13.8	&--	&--	&--	&--	&--	&--	&--	&--	&--	&--	&-- \cr
&SRR-FSD \cite{ref33} 	&14.7	&29.2	&13.5	&--	&--	&--	&--	&--	&--	&--	&--	&-- \cr
&N-PME \cite{ref34} 	&14.1	&26.5	&\underline{13.6}	&\textbf{6.9}	&13.3	&21.5	&18.6	&\textbf{31.4}	&\textbf{32.4}	&\textbf{14.2}	&30.7	&\textbf{46.6} \cr
\rowcolor{gray!30}
&DRL (normal)$ ^\ast $       	&14.6	&\underline{31.3}	&11.3	&4.8	&\underline{15.5}	&\underline{22.3}	&22.1	&28.7	&28.8	&10.8	&30.0	&40.4 \cr
\rowcolor{gray!30}
&DRL (residual)$ ^\ast $	&\underline{15.0}	&\textbf{31.7}	&11.8	&4.8	&\textbf{15.9}	&\textbf{23.1}	&\textbf{22.6}	&\underline{29.6}	&\underline{29.7}	&11.1	&\underline{31.0}	&\underline{41.3} \cr\hhline{==============}

\end{tabular}

\begin{tablenotes}
\footnotesize 
\item $ ^{\ddag} $   the reported results are reproduced using the official pre-training model and code. 
\item $ ^{\ast} $ the reported results are the average of ten random runs.
\end{tablenotes}
\end{threeparttable}

\end{table*}

\subsection{Benchmarks and Setups}
To make a fair comparison, our experimental setups on Pascal VOC 2007, 2012 and MS-COCO 2014 datasets are consistent with \cite{ref24}. Following the description in \cite{ref26}, three different schemes are utilized to take five classes as novel classes from the original Pascal VOC dataset (20 object categories in total) to perform few-shot detection. Each class has only $ K $  pictures to participate in the fine-tune training stage, $ K $=1,2,3,5,10. The remaining 15 classes are base classes, providing sufficient samples to participate in both the base and fine-tune training stages. MS-COCO dataset is a more challenging dataset, which contains 80 categories in total. Among them, 20 categories belonging to Pascal VOC are regarded as novel classes, and the other 60 categories are regarded as base classes. 

\subsection{Analysis of experimental results}

\subsubsection*{\bf Pascal VOC }
The experimental results are presented in Table I. It can be seen that the proposed model with two GCN structures (normal and residual ones) achieves the highest accuracy in almost all the setups with three different partitions of dataset and different number of novel class samples. Compared with State-Of-The-Art (SOTA) method FSDVE \cite{ref24}, our model shows constant improvement in term of $ AP_{50} $, except the 5-shot setting in novel set 2. 

Since different novel sets contain different novel classes, the performance varies from set to set. For example, in the Novel Set 1, the five novel classes are “bird, bus, cow, motorbike, sofa”. The biggest improvement is 7.2\% in the 3-shot setting of residual structure. However, in the 5-shot setting, the improvement is relatively minor (0.8\% for the residual structure), which leads to very close result in our model of 3-shot and 5-shot. Although it performs well when there are only limited samples, the performance does not scale linearly when the number of samples is increased from 3 to 5. The similar cases can be found in Novel Set 2. 

In Novel Set 2 with the novel classes "aeroplane, bottle, cow, horse, sofa", the $ AP_{50} $ from 3-shot to 5-shot decrease 0.3\% of residual structure. The best guess is that the proposed DRL model is less effective at handling small objects compared to medium and large ones, which can be observed in the MS-COCO experiments as well. Moreover, in the $ AP_{50} $ of each category shown in Table II, the smallest object “bottle” has the worst performance for both the proposed DRL and FSDVE. The main reason is that the support and query images of “bottle” usually contain a more complicated image background, such as a cluttered desktop. Noting that our DRL model still outperforms the compared method, thanks to the relationship between the features built by the dynamic GCN. Despite this, the proposed DRL model shows noticeable performance on 1, 2, 3-shot detection. In the 10-shot setting, the DRL gets improvement up to 7.0\% when each category gets stable representation. 

In contrast, the accuracy of DRL scales well with the sample numbers and constantly higher than the SOTA method FSDVE in Novel Set 3. The five novel classes are "boat, cat, motorbike, sheep, sofa", which all have moderate sizes.

\begin{table}[!t]
\renewcommand\arraystretch{1.2}
\caption{Comparisons of The Meta Loss and DRL (MS-COCO 10-Shot, \%)\label{tab:table4}}
\centering
\setlength\doublerulesep{0.5pt}
\begin{threeparttable}
\setlength{\tabcolsep}{3.6mm}
\begin{tabular}{cccccc}
\hhline{======}
Meta Loss &DRL &$ AP $  &$ AP_{50} $ &$  AR_1 $ &$ AR_{10} $  \cr\hline
&  &11.1	&26.3 &18.8 &24.8 \cr
$ \checkmark $  & &10.5	&25.5	&18.6	&23.8 \cr
&$ \checkmark $ &\textbf{11.9}	&\textbf{27.4}	&\textbf{19.7}	&\textbf{25.2} \cr
$ \checkmark $ &$ \checkmark $ &10.9	&26.1 &19.2 &24.7 \cr
\hhline{======}
\end{tabular}

\begin{tablenotes}
\footnotesize 
\item The reported results are obtained using DRL (normal).
\end{tablenotes}
\end{threeparttable}

\end{table}

\begin{table}[!t]
\renewcommand\arraystretch{1.2}
\caption{Base Training With/Without DRL on VOC Novel Set 1 (\%).\label{tab:table5}}
\centering
\setlength\doublerulesep{0.5pt}
\begin{threeparttable}
\setlength{\tabcolsep}{2.8mm}
\begin{tabular}{cccc|ccc}
\hhline{=======}
&\multicolumn{3}{c|}{Base (mAP)} &\multicolumn{3}{c}{Novel (mAP)} \cr Shots &1 &2 &10 &1 &2 &10 \cr\hline
\rule{0pt}{15pt}
\makecell[c]{Base training \\ Fine-tune}  &\textbf{62.3}	&\textbf{64.3}	&\textbf{68.5}	&\textbf{28.6}	&40.3	&59.4 \cr
\rule{0pt}{15pt}
\makecell[c]{Only \\ Fine-tune} &59.7	&62.8	&67.9	&28.0	&\textbf{40.5}	&\textbf{59.4} \cr
\hhline{=======}
\end{tabular}

\begin{tablenotes}
\footnotesize 
\item The reported results are obtained using DRL (residual). 
\end{tablenotes}
\end{threeparttable}

\end{table}

\subsubsection*{\bf MS-COCO}
The experimental results compared with the SOTA methods are shown in Table III. The blanks in the table indicate that neither the results were not reported in the corresponding papers nor the official codes haven’t been released. The standard MS-COCO evaluation protocol is followed, which includes mean Average Precision (AP) with different Intersection over Union (IoU) with ground truth, and AP with objects occupying areas of different sizes. At the same time, we also add mean Average Recall (AR) with diffident top proposals as a measure. So that we can evaluate the model performance more comprehensively. Compared with Pascal VOC, the improvement on MS-COCO is relatively small, because the images in MS-COCO is more complex and has more categories and samples. 

In general, compared with the latest methods in FSOD, our proposed DRL still maintains a comprehensive lead in the 10-shot setting. Although its $ AP $ is 0.3\% lower than FSCE \cite{ref31} in 30-shot setting, the results of DRL are still higher than the other SOTA methods. For instance, compared with the SOTA method FSDVE, in the 10-shot setting, our method has about 1.0\%-1.3\% improvement in both $ AP $ and $ AR $, which shows the effectiveness of the proposed DRL. Similarly, we found that the results on small objects are slightly lower (-0.2\%) than the SOTA method. However, there is a great improvement on large objects (over 2\% boost in  $ AP_L $ and $ AR_L $). This is actually an inherent problem in deep metric learning. We need to constrain the features by the distance between RoI and support features, in order to obtain more generalized features to help improve recognition performance. However, for small objects, the semantic information of the feature itself is not clear enough. Due to this, the constraint of feature distance may make it move in the wrong direction.  

In the 30-shot setting, the overall performance of DRL is still superior to the SOTA methods, with at least 0.2\%-0.4\% improvement in $ AP $ and $ AR $. At the same time, DRL is still not good at the detection of small objects. This also confirms the above analysis, small objects could cause semantic deviation, which may become more serious when the number of samples becomes more.

\begin{table}[!t]
\renewcommand\arraystretch{1.2}
\caption{Input Format of The Support Data on VOC Novel Set 1 (mAP, \%).\label{tab:table6}}
\centering
\setlength\doublerulesep{0.5pt}
\begin{threeparttable}
\setlength{\tabcolsep}{3.2mm}
\begin{tabular}{cccccc}
\hhline{======}
Shots &1 &2 &3 &5 &10  \cr\hline
\rule{0pt}{12pt}
image + mask &\textbf{28.1}	&\textbf{41.9}	&47.4	&\textbf{49.9}	&58.4 \cr
\rule{0pt}{12pt}
instance	&28.0	&40.5	&\textbf{49.4}	&\textbf{49.9}	&\textbf{59.4} \cr
\hhline{======}

\end{tabular}

\begin{tablenotes}
\footnotesize 
\item The reported results are obtained using DRL (residual). 
\end{tablenotes}
\end{threeparttable}

\end{table}

\subsection{Ablation studies}
\subsubsection*{\bf Two structures of GCN}
The results of two proposed GCN structures, namely normal and residual GCN, have been presented in Tables I and III.  It can be seen that, in 1 or 2-shot settings in PASCAL VOC and 10-shot setting in MS-COCO, the performance of normal GCN is higher than that of the residual GCN. With such few samples, support images cannot provide stable class representation. In the same time, support features are used to update the prediction probability more frequently in the residual GCN. Inaccurate class representation will make the error further expand, and normal structure can alleviate this situation.

As the number of samples increase (5-10 shots in PASCAL VOC or 30 shots in MS-COCO), the performance of the residual GCN has been noticeably improved. This difference shows the number of anchor nodes is important for the proposed dynamic relevance learning. It is because the amount of available support features indicates how many direct connections with strong relevance between nodes are available. Moreover, the residual structure may be able to alleviate the inaccurate influence between drift nodes (RoI features). 

\subsubsection*{\bf Influence of the meta loss}
Meta loss was first proposed in Meta R-CNN \cite{ref22}, which is a simple but effective way. The Cross Entropy function is used to calculate the loss of image classification in support set, trying to expand the distance between inter-classes, so that different categories have diverse effects on query images. However, in our experiments, we find that meta loss is not suitable for the fine-tune training stage as stated in Section III.D Training strategy. 

As shown in Table IV, with and without the meta loss, the $ AP $ of novel classes has about 0.6\% difference. The combination of the meta loss with the proposed DRL will also degrade the performance. The best guess is that the meta loss forced classification constraint on a certain class, instead of directly adding constraints to the distance between features. When there are enough samples, this method can make the features of different categories more diversity. But in case of very few samples, which may make the representation of the class deviate and thus cannot represent the class well. Our DRL still shows good performance even in the case of insufficient samples.

\begin{figure}[!t]
\centering
\includegraphics[width=3.5in]{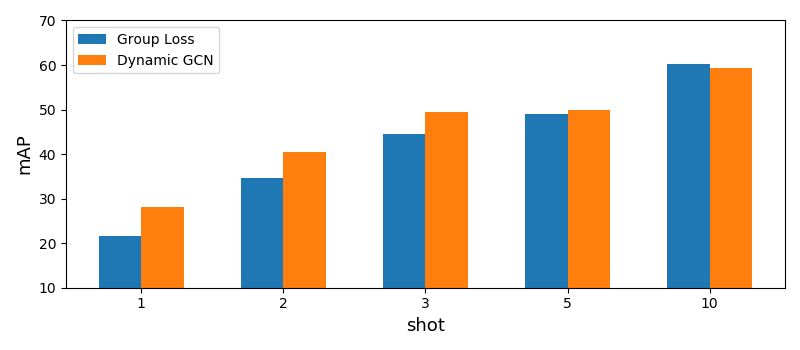}
\caption{The comparisons between Group Loss and dynamic GCN on the VOC Novel Set 1. The results are obtained using DRL (residual).}
\label{fig_5}
\end{figure}

\begin{figure}[!t]
\centering
\includegraphics[width=3.5in]{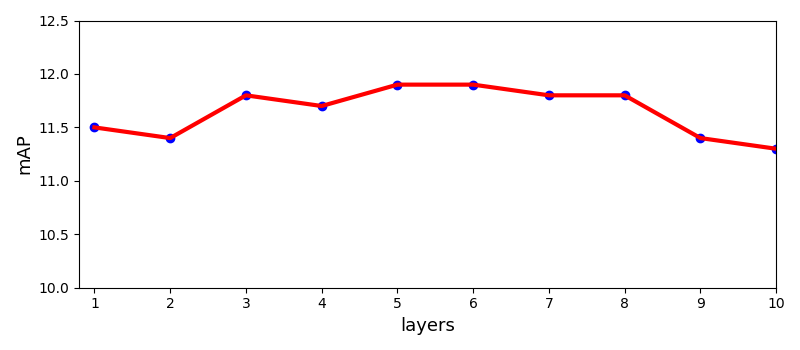}
\caption{The effect of different depths of GCN in MS-COCO 10-shot setting. The results are obtained using DRL (normal).}
\label{fig_5}
\end{figure}

\subsubsection*{\bf Comparisons between the Group Loss and Dynamic GCN}
The proposed DRL model is inspired by Group Loss. It iterates the initial class probabilities $ \{p_j\}_{j=1}^{n_{roi}} $ of all RoI in a fixed way through the feature distance $ S $ between samples,

\begin{equation}
\label{eq13}
\pi_{m,c} = \sum_{j=1}^N S_{m,j}p_{j,c},
\end{equation}
where $ \pi_{m,c} $ can be regarded as the confidence that RoI feature $ r_m $ belongs to class $ c $, which is determined by the similarity $ S_{m,j} $  between $ m $-th and $ j $-th samples as well as the probability of the j-th sample for class $ c $. Then the enhanced probability of $ r_{m,c} $ for class $ c $ can be obtained by multiplying the confidence $ \pi_{m,c} $ and initial probability $ p_{m,c} $   with a normalization step,
\begin{equation}
\label{eq14}
r_{m,c}(t+1) = \frac{r_{m,c}(t)\pi_{m,c}(t)}{\sum_\mu r_{m,\mu}(t)\pi_{m,\mu}(t)},
\end{equation}
where $ t $ indicates $ t $-th iteration, and $ r_{m,c}(0) = p_{m,c} $. 
 
The comparisons are demonstrated in Fig. 5. The results of the group loss are inferior to DRL in 1, 2, 3 and 5-shot detection. In the case 10-shot, group loss and DRL obtain very close results. With 10 or more samples per class, the model could get more stable category representation. Therefore, the fixed iteration method can also establish reliable connection between support data and query images, although the number of iterations still need to be determined manually.

\begin{table}[!t]
\renewcommand\arraystretch{1.5}
\setlength{\tabcolsep}{1.5mm}
\caption{Comparisons of Different Similarity Metrics (MS-COCO 10-Shot, \%).\label{tab:table7}}
\centering
\setlength\doublerulesep{0.5pt}
\begin{threeparttable}

\begin{tabular}{c|cccccc}
\hhline{=======}
Methods &$ AP $  &$ AP_S $ &$ AP_L $ &$ AR_{100} $ &$ AR_S $ &$ AR_L $ \cr\hline
Cosine Similarity	&10.2	&3.3	&14.6	&23.7	&8.0	&31.0 \cr
Euclidean Distance	&10.8	&3.6	&14.9	&24.3	&8.7	&31.3 \cr
Gaussian Kernel	&11.0	&\textbf{4.0}	&15.8	&24.0	&8.1	&31.8 \cr
Neural Network	&11.1	&3.9	&17.1	&\textbf{25.7}	&\textbf{9.2}	&\textbf{36.5} \cr
Pearson’s Correlation	&\textbf{11.9}	&3.8	&\textbf{17.7}	&25.3	&8.6	&35.5 \cr

\hhline{=======}

\end{tabular}

\begin{tablenotes}
\footnotesize 
\item The reported results are obtained using DRL (residual). 
\end{tablenotes}
\end{threeparttable}

\end{table}

\subsubsection*{\bf Base training with DRL}
To show the effectiveness of DRL in different training stages, the results with and without DRL in base training stage are given in Table V. It can be seen that using DRL solely in the fine-tune stage can produce good enough results for the novel classes. But for base classes, utilizing DRL can further push the $ AP $ up of 1.5\%-2.6\%. It also reflects that the proposed DRL can help p re-trained model to quickly adapt to new tasks. 

\subsubsection*{\bf Different input format of support images}
To get clearer object information, there are generally two input formats for support images. One contains only the instance, i.e. the object part in the image. The other provides the whole 3-channel image with a binary mask channel, while the instances and background are marked as 1 and 0, respectively. The influence of the input format has been discussed in MetaYOLO, and the conclusion is that adding binary mask will bring about 2\% $ AP $ improvement. In this paper, we show different results with the proposed DRL, As given in Table VI, the whole image + mask is better in 1 and 2-shot settings, while its performance is inferior to the instance in 3 and 10-shot settings. 

The reason of such difference may be that the feature extractors of support and query images in MetaYOLO do not share parameters. Hence, the addition information in the mask channel may help the feature extraction of support images, which leads to a better performance. In contrast, the feature extractors of the proposed DRL model have the same parameters, while the dynamic GCN introduces further information exchange.

\begin{figure*}[!t]
\centering
\includegraphics[width=7in]{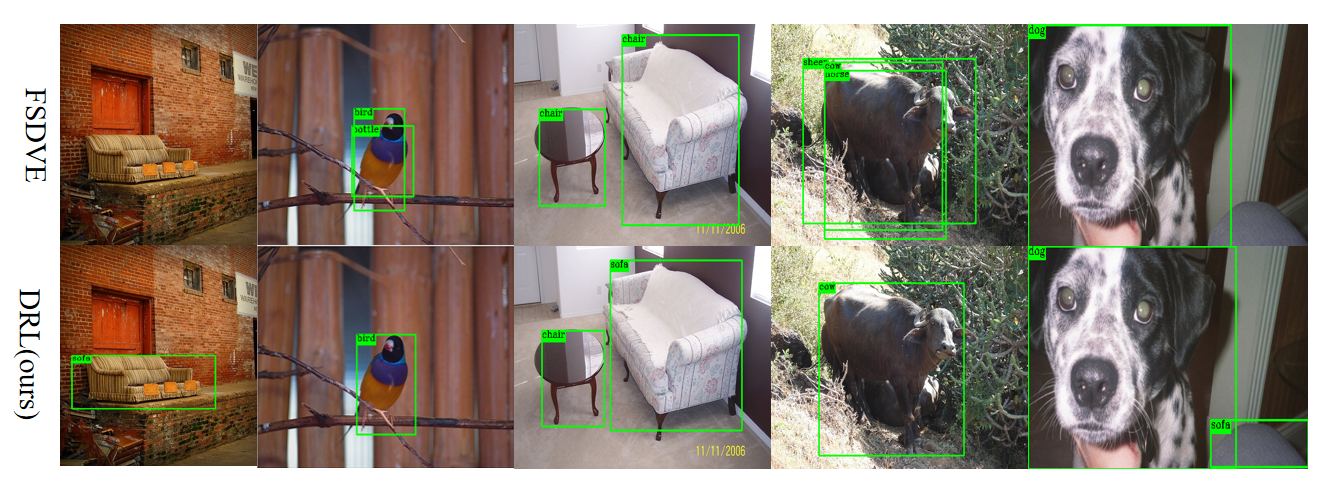}
\caption{The visualization of novel classes objects on VOC Novel Set 1 detected by FSDVE and our DRL, the structure of GCN is Residual.}
\label{fig_7}
\end{figure*}

\begin{figure*}[!t]
\centering
\subfloat[]{\includegraphics[width=2.3in]{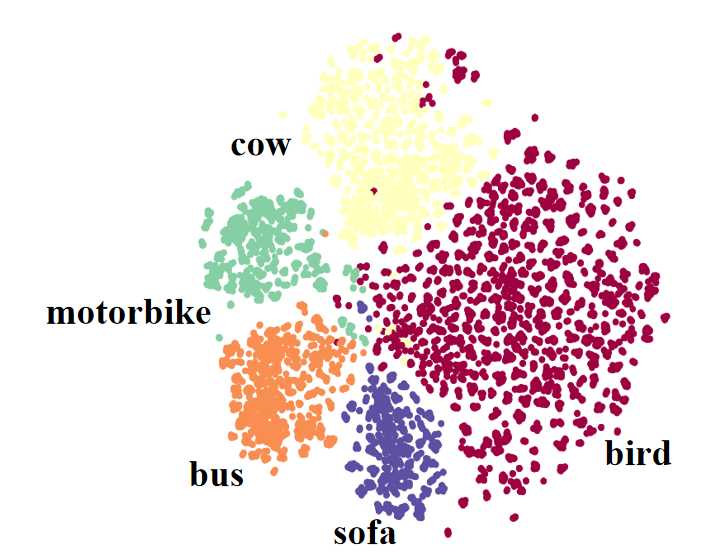}%
\label{fig_novelset1}}
\hfil
\subfloat[]{\includegraphics[width=2.3in]{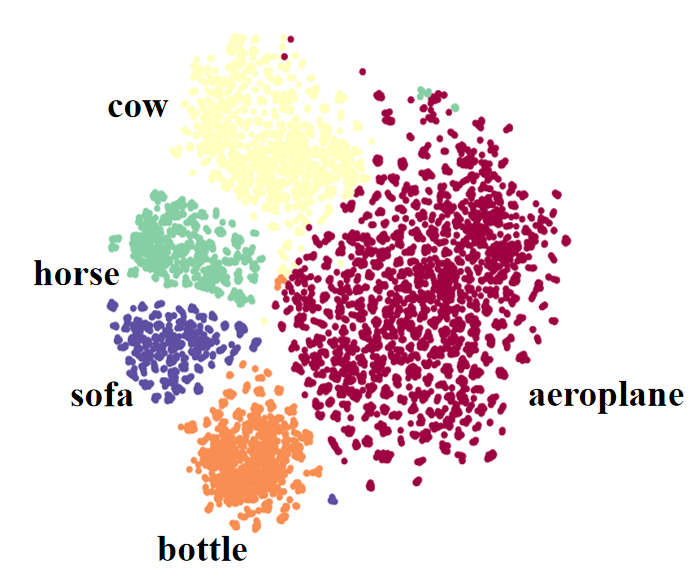}%
\label{fig_novelset2}}
\hfil
\subfloat[]{\includegraphics[width=2.3in]{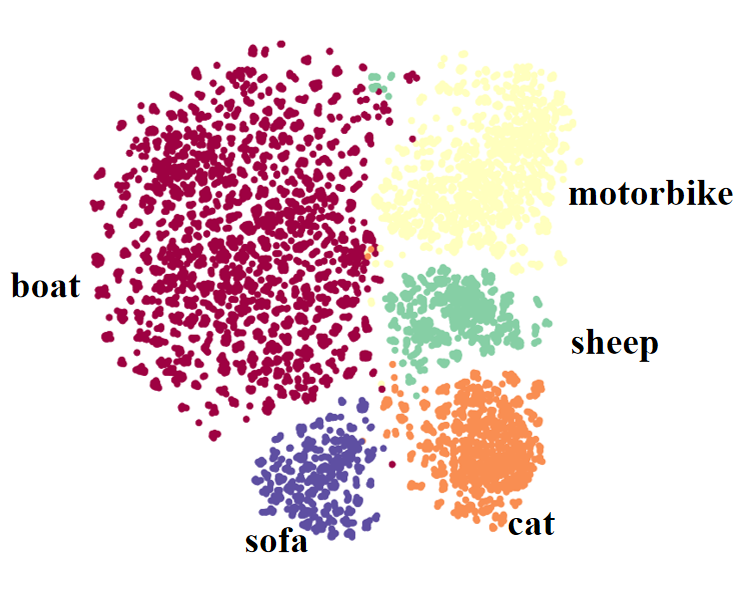}%
\label{fig_novelset3}}
\caption{The t-SNE of RoI features of novel classes on the (a) VOC Novel Set 1, (b) VOC Novel Set 2 and (c) VOC Novel Set 3. The results are obtained using DRL (residual) in 10-shot setting.  }
\end{figure*}

\subsubsection*{\bf Different number of layers in dynamic GCN}
The number of model layers is always a vital hyper-parameter.  The network structure of traditional GCNs is much shallower than CNNs. In the proposed DRL, the dynamic GCN is employed which is more complicate than the static one. Thus, the effect of different depths for normal dynamic GCN structure is given in Fig. 6. It can be seen that the performance improves slightly when the number of layers increasing from 1 to 6. After the number exceed 7, the performance starts declining gradually. In other words, the performance is not very sensitive to the number of layers for the proposed DRL model.

It is worth noting that the additional parameters and computational cost of DRL are ignorable compared to other baseline models. Specifically, the number of parameters and MACs of DRL with a 10-layer GCN for MS-COCO dataset are only approximately 64.0K and 25.6K, respectively. In contrast, FSDVE requires about 50M parameters and 45G MACs. Moreover, DRL is only used in training stage, which does not increase the burden of inference.

\subsubsection*{\bf Different similarity metrics}
The Pearson’s correlation coefficient used in this work is inspired by the previous work of group loss, since it provides scaling and translation invariance by data standardization. Meanwhile, other similarity metrics, such as the cosine similarity, Euclidean distance, gaussian kernel function or neural network mentioned in TRPN, can be the alternative to calculate the relation matrix $ S $.

To analyze the influence of different similarity metrics, the comparison on MS-COCO dataset (10-shot) is presented in Table VII. It can be seen that Pearson’s correlation coefficient obtained the highest AP compared with other metrics, which is 0.8\% higher than the second highest method (neural network). In contrast, the AP and AR results of cosine similarity are the lowest indicating that it is not suitable for DRL. Normally, the performance of cosine similarity should be better than Euclidean distance in most cases. However, Euclidean distance outperforms cosine similarity in our experiments, which has been also reported in Prototypical Network [12]. They conjecture this may due to cosine distance not being a Bregman divergence, which leads to inferior cluster representatives. It is also worth noting that the neural network method achieves the highest AR, which shows the capability of adaptive learned metric function to capture the difference between foreground and background features.

\subsection{Visualization of the detection results}
To demonstrate the superiority of DRL more intuitively, we visualize several detection results of five novel classes on VOC Novel Set 1 in Fig. 7. In the first and fifth pictures, the appearance of the sofa is like the background, thus FSDVE mistakenly classifies it as the background. In the fine-tuning phase, our DRL uses the support feature as a reference to adjust the sample features, resulting a greater distance between the background and object features of RoI, which leads to better results. The second to the fourth contain similar mistakes for FSDVE. Although the object is correctly located and classified, the background is mistakenly classified as the foreground object. In contrast, our model can alleviate such problem effectively. 

\subsection{Visualization of RoI features}
RoI features learned by DRL are visualized using t-SNE [65] in Fig. 8, which can provide a more intuitively observation of the relationship between different classes of features. It can be seen that the RoI features of the same category are well clustered, and there are obvious boundaries between different categories. This shows that DRL is effective to constrain the learning of RoI features by using the support features as guidance information. Meanwhile, it is also capable to learn the correct semantic information when there are only 10 training samples in each category. 

\section{Conclusion}
In this paper, to improve the performance of few-shot object detection, we have proposed a dynamic relevance learning model, which can guide the learning of category representation by utilizing the relationship between the support and query images. Two types of dynamic GCNs have been constructed and tested, which take the embeddings of the sample as the graph node and update the information by graph convolution. The experimental results and ablation study show that the DRL is effective in FSOD, due to its ability to learn more generalized features.

\vfill

\end{document}